\documentclass[colorlinks=true, linkcolor=black, urlcolor=blue, citecolor=blue, letterpaper, 10 pt, conference]{ieeeconf}

\IEEEoverridecommandlockouts  

\usepackage{color}
\usepackage{array}

\usepackage{graphicx} 
\usepackage{hyperref}

\usepackage{mathptmx} 
\usepackage{times} 
\usepackage{amsmath} 
\usepackage{amssymb}  
\usepackage{color}
\usepackage{bm}
\usepackage{rotating}
\usepackage{subfigure}
\usepackage{tabularx}
\usepackage{colortbl}
\usepackage{hhline}
\usepackage{multirow}
\usepackage{verbatim}
\usepackage{cite}
\usepackage{siunitx}
\usepackage{graphicx}
\usepackage{duckuments}
\usepackage{soul,xcolor}
\setstcolor{red}
\usepackage[ruled, lined, linesnumbered, commentsnumbered, longend]{algorithm2e}
\usepackage[capitalise,noabbrev]{cleveref}
\usepackage{amsfonts}
\usepackage{siunitx}
\usepackage{booktabs}

\usepackage{multirow}
\usepackage[flushleft]{threeparttable}
\usepackage[acronym]{glossaries}
\glsdisablehyper

\newacronym{dof}{DoF}{Degree of Freedoms}
\newacronym{tof}{ToF}{Time-of-Flight}
\newacronym{hri}{HRI}{Human Robot Interaction}
\newacronym{fov}{FoV}{Field of View}
\newacronym{slam}{SLAM}{Simultaneous Localization and Mapping}
\newacronym{pf}{PF}{Particle Filter}
\newacronym{pc}{PC}{Point Cloud}
\newacronym{mcl}{MCL}{Monte Carlo Localization}

\begin{document}
	
\title{\LARGE \bf
    Tiny LiDARs for Manipulator Self-Awareness:  Sensor Characterization and Initial Localization Experiments
 }


\author{\hspace{0.2in} Giammarco Caroleo$^{1*}$, Alessandro Albini$^{1}$, Daniele De Martini$^{1}$, Timothy D. Barfoot$^{2}$, Perla Maiolino$^{1}$
\thanks{$^{1}$ are with the Oxford Robotics Institute, University of Oxford, UK.}
\thanks{$^{2}$ is with the University of Toronto Robotics Institute, Canada.}
\thanks{$^{*}$Corresponding author. Please contact at the email address \href{mailto:giammarco@robots.ox.ac.uk}\texttt{giammarco@robots.ox.ac.uk}}
\thanks{This work was supported by the SESTOSENSO project (HORIZON EUROPE Research and Innovation Actions under GA number 101070310).}
}

\maketitle


\newpage
\begin{abstract}
For several tasks, ranging from manipulation to inspection, it is beneficial for robots to localize a target object in their surroundings. In this paper, we propose an approach that utilizes coarse point clouds obtained from miniaturized VL53L5CX Time-of-Flight (ToF) sensors (\textit{tiny LiDARs})
to localize a target object in the robot's workspace. We first conduct an experimental campaign to calibrate the dependency of sensor readings on relative range and orientation to targets. We then propose a probabilistic sensor model, which we validate in an object pose estimation task using a Particle Filter (PF). The results show that the proposed sensor model improves the performance of the localization of the target object with respect to two baselines: one that assumes measurements are free from uncertainty and one in which the confidence is provided by the sensor datasheet.
\end{abstract}

\section{Introduction}\label{sec:intro}

Collaborative robots are required to operate in highly dynamic environments while ensuring safety.
Consequently, localization is essential for them to maintain awareness of their changing surroundings and adapt their actions accordingly.
In particular, LiDAR systems have been widely utilized in the literature for tasks such as object location and \gls{slam}, thanks to their reliability and ability to provide high-density measurements~\cite{zou2022,khan2021}.

\begin{figure}[t!]
    \centering
    \includegraphics[width=0.9\columnwidth]{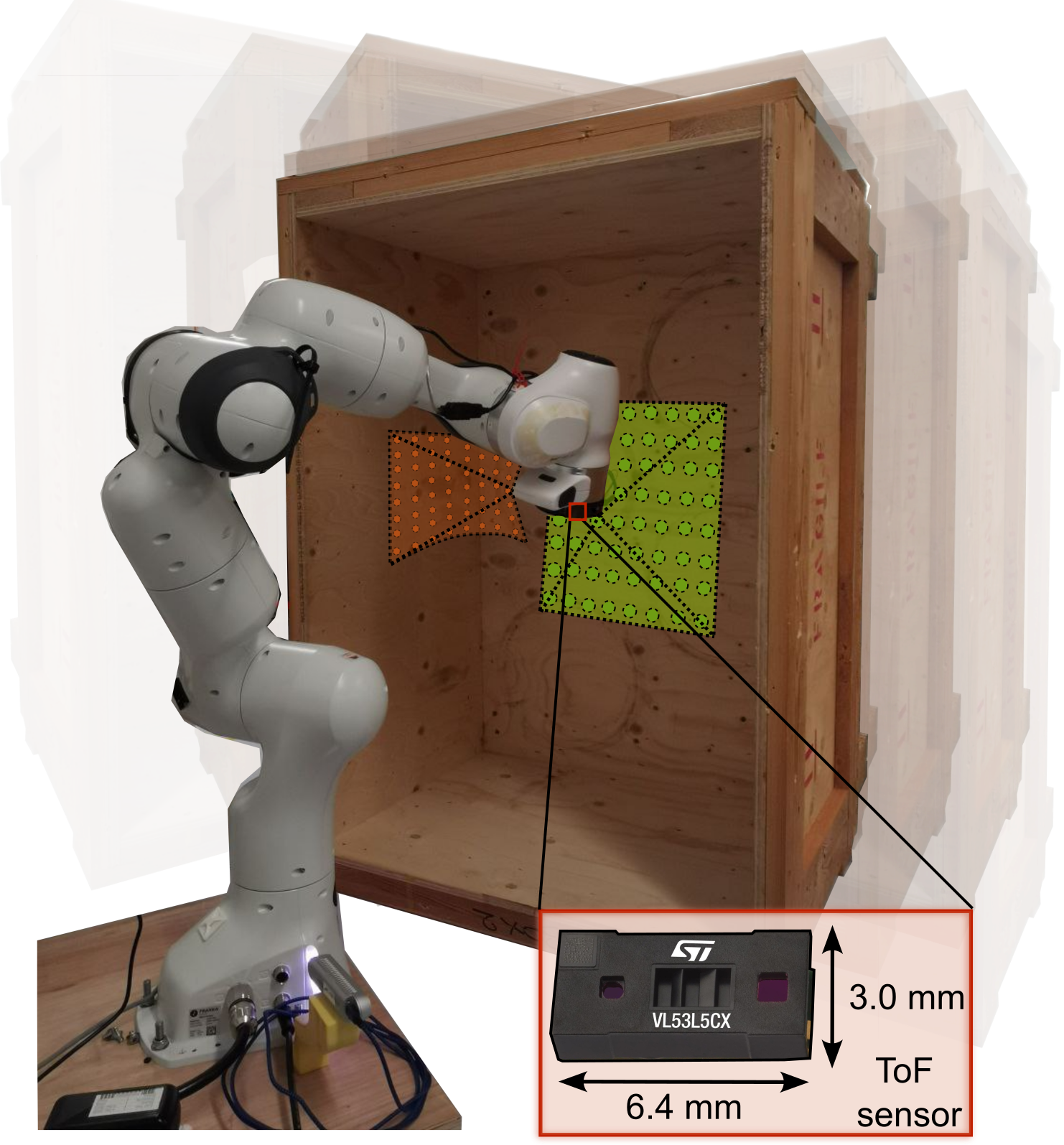}
    \caption{Robot estimating the pose of the crate using a Particle Filter based on measurements collected with miniaturized \acrshort{tof} sensors. The faded crates illustrate the estimated poses associated with particles. Each sensor provides an 8×8 depth map of the surface falling into the \acrshort{fov}, which can be converted into a small Point Cloud. A picture of the sensor used in this paper and its dimensions are also reported.}
    \label{fig:intro}
\end{figure}

Recently, miniaturized LiDARs based on \gls{tof} technology have garnered significant interest as a cost-effective (£6.50 each) and lightweight (approximately \SI{1}{\gram}) alternative to traditional LiDAR systems~\cite{callenberg2021low}. These sensors have so far mainly been employed in obstacle and collision avoidance, pre-touch sensing, in-hand manipulation, object detection, and gesture recognition~\cite{ding2019proximity, ding2020collision, caroleo, al2020towards, yang2017pre, sasaki2018robotic, koyama2018high, ruget2022pixels2pose}. Their characteristics, though, along with their extremely compact form factor, make them particularly suitable for distributed integration into the robot body, facilitating medium-range distance sensing (typically up to a few meters), enhancing spatial awareness~\cite{hughes2018robotic, tsuji2021sensor, caroleo2025soft} and extending their range of applications to object localization and \gls{slam}.
The advantage of the ease of integration comes at the cost of significantly lower spatial resolution of the sensor measurements compared to traditional LiDAR (most recent models provide an 8×8 depth map). In addition, \gls{tof} sensors typically feature a pyramidal \gls{fov}, which results in increasingly coarse spatial resolution the higher the distance to the target  (see \cref{fig:intro}).
Their use with existing state-of-the-art algorithms performing \gls{slam} or localization presents idiosyncratic challenges, as these algorithms are designed to work with much higher-density point clouds of the surrounding environment. 

Work has been done to address this drawback by pairing the \gls{tof} with other sensors, as in \cite{liu2023multi} with a monocular camera and in \cite{ruget2022pixels2pose} with an RGB-D camera. In this work, we present an approach for localizing large objects in the robot's workspace. Unlike the aforementioned works, our method relies solely on using \gls{tof} sensors.
The sensor used in this work provides a small 8×8 depth map as output and is capable of sensing the relative distance from objects up to \SI{4}{\meter}. The localization is achieved by applying a \gls{pf} algorithm to a probabilistic beam model of the sensor. We draw inspiration from~\cite{lancaster2022optical}.
Their work focused on 2D in-hand localization of a small object at a very short distance;  therefore, compared to our case, their sensor measurements were less affected by uncertainties related to distance from the target.
We decided to deal with such uncertainties by performing a systematic sensor characterization to (i) correct the raw sensors measurements and (ii) build a more accurate probabilistic sensor model that accounts for the observed noise.

In summary, this work tackles the problem of localizing an object in the workspace of a robot manipulator equipped solely with \gls{tof} sensors. The probabilistic model, enhanced by our extensive sensor characterization, has been validated through comparative experiments to quantitatively analyze the improvement in the localization accuracy with respect to two baseline methods -- the first relies on information provided by the sensor datasheet, while the second assumes measurements are free from uncertainty.

This work is organized as follows.
We first introduce the notation and scope of the work in \cref{sec:methodology} and describe the characterization process in \cref{sec:calibration}.
\Cref{sec:validation} outlines our experimental setup and \cref{sec:results} details our results. Conclusion follows. 
\section{\gls{tof}-Based Localization}\label{sec:methodology}

We aim to determine the position of an object in the surroundings of a robotic manipulator using coarse and noisy measurements. 
To this end, we accommodate sensor noise by identifying a probabilistic sensor model that we then exploit in a \gls{mcl} algorithm using a \gls{pf}.
In this section, we first detail the sensor model for the localization task and then briefly describe the \gls{mcl} algorithm\footnote{We refer the reader to~\cite{THRUN200199} for a more exhaustive treatment on localization methods and stochastic sensor modeling.}.

\subsection{Probabilistic Sensor Model}

The miniaturized \gls{tof} sensor utilizes $K$ separate beams that measure the environment at different fixed directions, resulting in a detection volume.
As a result, at the sampling time $t$, the sensor produces $K$ distance measurements $z^k_t \in \mathbb{R}$ in the direction of the $k$-th beam.
We collect the measurements at time $t$ in the set \(Z_t =\lbrace z_t^1, \dots, z_t^K \rbrace \).

The inaccuracy over \(z_t^k\) is conditioned on the \gls{tof} sensor pose with respect to a target object \(o_t\), and on the environment itself.
Since the sensors are rigidly attached to the robot, their state is intertwined with the robot state \(r_t \in SE(3) \), corresponding to the Cartesian pose of the frame to which the \gls{tof} is rigidly connected.
Accordingly, we define the sensor model \( p(z_t^k \mid r_t, o_t) \) to represent the conditional probability of the observation \( z_t^k\) given the state \( r_t\) and the object \(o_t\). 

Given the nature of the \gls{tof} sensors, the model \( p(z_t^k \mid r_t, o_t) \) mirrors the beam range finder model~\cite{THRUN200199} that is widely used for range sensors and contemplates the effect that noise and unexpected objects have on range estimation. 
In our case, we want the model to mainly handle the sensor noise.
Indeed, the robot does not interact with the environment, i.e., has no effect on \(o_t\), and the object is supposed to be fixed over time. 
For this reason, in the following, we will refer to the object without a subscript.

We can then express the sensor noise as a Gaussian distribution centered at the \textit{true} range value \(z_t^{k*}\) and with variance $\sigma^2$. 
The true range value can be computed with a ray-casting technique given the pose \(r_t\) of the robot. Conversely, the variance is estimated by characterizing the sensor. 
Therefore, we have \( p(z_t^k \mid r_t, o) \sim \mathcal{N}(z_t^{k*}, \sigma^2) \) and~\(p(Z_t \mid r_t, o) = \prod_{k=1}^{K} p(z_t^k \mid r_t, o)\) as we assume the likelihood of the individual beam to be independent from the others as is commonly done in the literature~\cite{THRUN200199}.

\subsection{Monte Carlo Localization}

The probabilistic model described above is integrated into the \gls{mcl} algorithm to update the belief function of the object's pose.
To describe the object pose $x_t$, we define a set of \(M\) particles $X_t = \lbrace x_t^1, \dots, x_t^M \rbrace$, with the \(x_t^{j} \in SE(3) \) representing the pose of the object in the Cartesian space. 

In our case, the belief function \(\text{bel}(x_t) \) given by the particles is formally defined as
\begin{equation}
\text{bel}(x_t) = p(x_t \mid Z_{1:t}),
\end{equation}
where \( Z_{1:t} \) denotes the sequence of sensor observations up to time \( t \). 
The belief is updated according to the weights the \gls{mcl} assigns to each particle based on the measurements~\cite{THRUN200199}. To this end, for each particle, the probabilistic sensor model is used to estimate the likelihood of the observations given the object pose associated with the particle. In fact, for a given \(x_t^j\), knowing the pose of the robot $r_t$, it is possible to estimate the true measurements \(z_t^{k*}\) and compute \(p(Z_t \mid r_t, o)\). The weights computed by the \gls{mcl} play a role in the resampling phase during which new particles are drawn. 

\section{Sensor characterization}\label{sec:calibration}
A sensor model explicitly represents measurement noise through a conditional probability distribution. In addition, it is important to account for other sources of errors that can affect sensor readings.
For \gls{tof} sensors, like other range sensors~\cite{cooper2018range, kneip2009characterization}, various factors can introduce errors in measurements. These factors include the distance to the target, the relative orientation between the sensor and the target, ambient light conditions, environmental temperature, and the physical properties of the target, such as its color and material.
Nonetheless, characterizing the sensor for each of these sources is non-trivial. As our use-case consists of an indoor object localization, we characterized the sensor only considering its distance from the target and its orientation. 
Being an initial study, we designed our experiments so that lighting conditions, environmental temperature, and material properties do not vary.  

In this section, we first describe the experimental setup used for the characterization. Then, details on the experimental procedure follow with a description of how the obtained results are integrated into the probabilistic sensor model. 

\subsection{Experimental Setup for \gls{tof} Characterization }\label{subsec:calibration_setup}
\begin{figure}[]
    \centering
    \includegraphics[width=0.9\columnwidth]{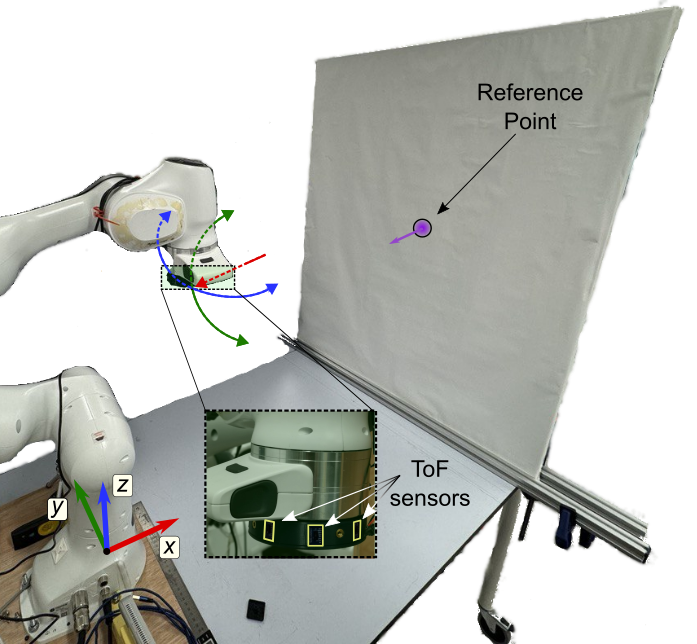}
    \caption{A board is wrapped with a white vinyl film and placed in front of the robot at a fixed distance. The robot is commanded to move: in the $x$ direction in the \SIrange{20}{800}{\milli\metre} range as highlighted with the red arrow; around the $z$-axis along a circular segment path of \SI{50}{\degree} by keeping a \SI{200}{\milli\metre} distance from the reference point as indicated with the blue arc; around the $y$-axis along a circular segment path of \SI{50}{\degree} by keeping a \SI{200}{\milli\metre} distance from the reference point as indicated with the green arc.}
    \label{fig:calibration setup}
\end{figure}
This work uses VL53L5CX \gls{tof} sensors from STMicroElectronics\footnote{Sensor datasheet available at:
\url{https://www.st.com/resource/en/datasheet/vl53l5cx.pdf} [2025].}, featuring a square \gls{fov} with a diagonal of \SI{65}{\degree}, a measurement range of \SI{20}{\milli\metre} to \SI{4}{\metre} and an acquisition frequency of \SI{15}{\hertz}. 
Each \gls{tof} sensor measures distance along 64 individual beams evenly distributed within the detection volume and provides an 8×8 matrix, with each element representing the distance in the direction of a specific beam. Assuming the direction of the beams is known from the specifications, distance measurements can be converted into a small \gls{pc} data structure. 

For the characterization, three sensors were mounted on the end-effector of a Franka Robotics Panda robotic arm using a 3D-printed support with a circular shape; hence the precise relative placement between the robot and the support, as well as between the support and the sensors, was given by the CAD model.
A flat board wrapped in a white vinyl film was placed orthogonal to the $x$-axis of the robot base frame as shown in~\cref{fig:calibration setup}.
The distance from the robot base was measured using a measuring tape as in~\cite{kneip2009characterization}, with a resolution of $10^{-3}$ m; while a proper orientation of the target was ensured by using a protractor with a resolution of \SI{0.1}{\degree}.

The robot was commanded to move its end-effector to different distances from the target, while also varying its orientation. All the readings are expressed with respect to the common reference frame associated with the robot base (see~\cref{fig:calibration setup}). Similar to \cite{lancaster2022optical}, the robot pose is assumed to be known with no uncertainties. 
Given the known position of the target board with respect to the robot and knowing the exact position of the \gls{tof} sensor with respect to the end-effector,
we computed the distance between the reference point on the board (see \cref{fig:calibration setup}) and the sensor, i.e., the true range $z^*$. 
In all the experiments, the reference point was kept close to the center of the board, so that all \gls{tof} measurements intersected the board in each test position.

The experimental procedure described in the remainder of the section was performed for the three \gls{tof} sensors, one at a time.

\subsection{Range Characterization}\label{subsec:range}

\begin{table}[]
\centering
\caption{Standard Deviation of Readings at Different Ranges}
\begin{tabular}{@{}ccccc@{}}
\toprule
\textbf{Range (mm)} & \textbf{Standard Deviation (\%)} & \textbf{Trend} \\ \midrule
20                 & 40.0                           & Maximum deviation \\
25                 & 1.4                            & Sudden drop from 20 mm \\
60                 & 1.2                            & Linear decrease from 25 mm \\
100         & 0.6 &Linear decrease from 60 mm \\
100 - 800          & 0.6                            & Constant deviation \\
\bottomrule
\end{tabular}\label{stddevrange}
\end{table}

\begin{figure*}[h]
    \centering
    \includegraphics[width=0.9\textwidth]{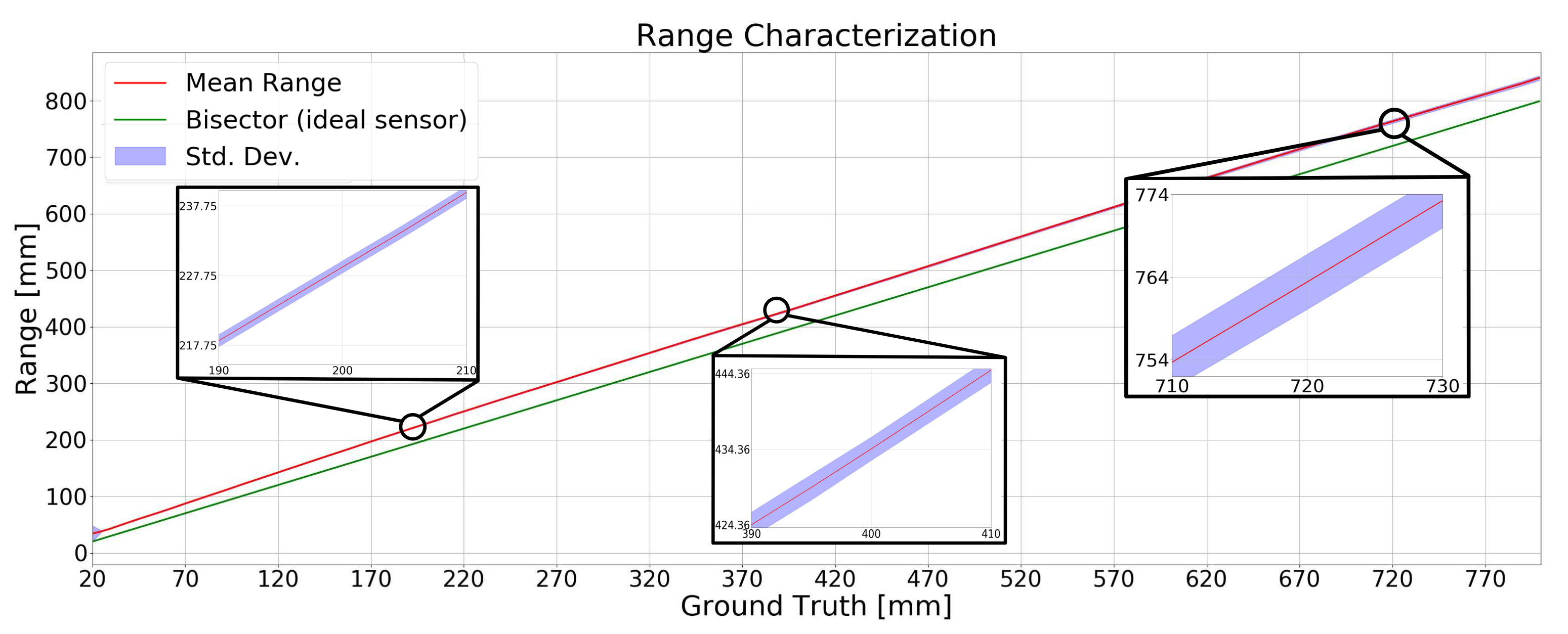}
    \caption{The plot presents the estimated range along with the deviation and zoom-ins to highlight how the deviation changes at different ranges. The green bisector line represents an ideal sensor with perfect accuracy, illustrating the deviation of the real sensor from ideal behavior.} 
    \label{fig: range calibration}
\end{figure*}
For the range characterization, the robot was positioned so that the normal of the \gls{tof} sensor of interest was orthogonal to the target at a distance of \SI{20}{\milli\meter}. 
As indicated in the datasheet, below this threshold the sensor does not provide reliable data. Starting from this pose, the robot was commanded to move backward in steps of fixed length up to a distance of \SI{800}{\milli\meter}. For the first \SI{200}{\milli\meter} the robot moved in \SI{5}{\milli\meter} increments and then in \SI{10}{\milli\meter} ones. 

We acquired 100 sensor samples at each commanded pose, resulting in 100 sets of measurements, each composed of 64 values corresponding to the number of beams.
We averaged each set -- compensated for beam angle by the sensor's firmware -- to estimate the trend of the average $\bar{z}$ range and its standard deviation in the $x$ direction (see~\cref{fig:calibration setup}) over the 100 samples at different distances. 
\Cref{fig: range calibration} shows the trend of the mean recorded at different ranges against the ground truth distances and the mean deviation recorded at each pose. 
The measurements are linearly correlated with the ground truth values with a deviation and an offset with respect to the ideal sensor measures (i.e., the precise measurements referring to the robot's pose without noise and error), represented by the green bisector line.
Therefore, we interpolated the data to find this correlation and found that the true range $z^*$ (in mm) can be linked to the measured range $\bar{z}$  according to the following equation:
\begin{equation}\label{eq:range}
    z^* = 0.963\times\bar{z} - 18.15~[mm].
\end{equation}
The discrepancy between measured data and the fitting line is on average below $1\%$ with a maximum of $17.5\%$ observed at \SI{20}{\milli\meter}.
It is worth noting that in the \SIrange{100}{800}{\milli\meter} range, the standard deviation associated with the readings varies constantly by 0.6\% of the measured range. Conversely, a maximum 40\% deviation is observed at \SI{20}{\milli\metre}. It then linearly drops to 1.4\% at \SI{25}{\milli\metre} and to 1.2\% at \SI{60}{\milli\metre}; it eventually reaches 0.6\% at \SI{100}{\milli\metre} in a linear fashion. We report these results in~\cref{stddevrange}.
The obtained results were repeatable across tests performed for the 3 different sensors. Therefore, there is no need for a tailored function for each to model the range error and estimate the noise distribution.

Notably, these results are significantly different from the sensor datasheet. In particular, the registered offset as well as the deviation from an ideal sensor measuring the true range are not mentioned. Furthermore, the reported deviation at different distances is generally greater or less informative. As an example, in the \SIrange{20}{200}{\milli\meter} range a deviation of \SI{15}{\milli\meter} is reported, which implies a deviation that varies from 75\% to 0.075\%. 

The results reported herein are used in a twofold way in the probabilistic model -- firstly, raw \gls{tof} measures are corrected based on the measured distance; secondly, the $\sigma$ for the given distance is assigned by \cref{stddevrange} via linear interpolation.

\subsection{Orientation Characterization}\label{subsec:orientation}

\begin{figure}[]
    \centering
    \includegraphics[width=0.9\columnwidth]{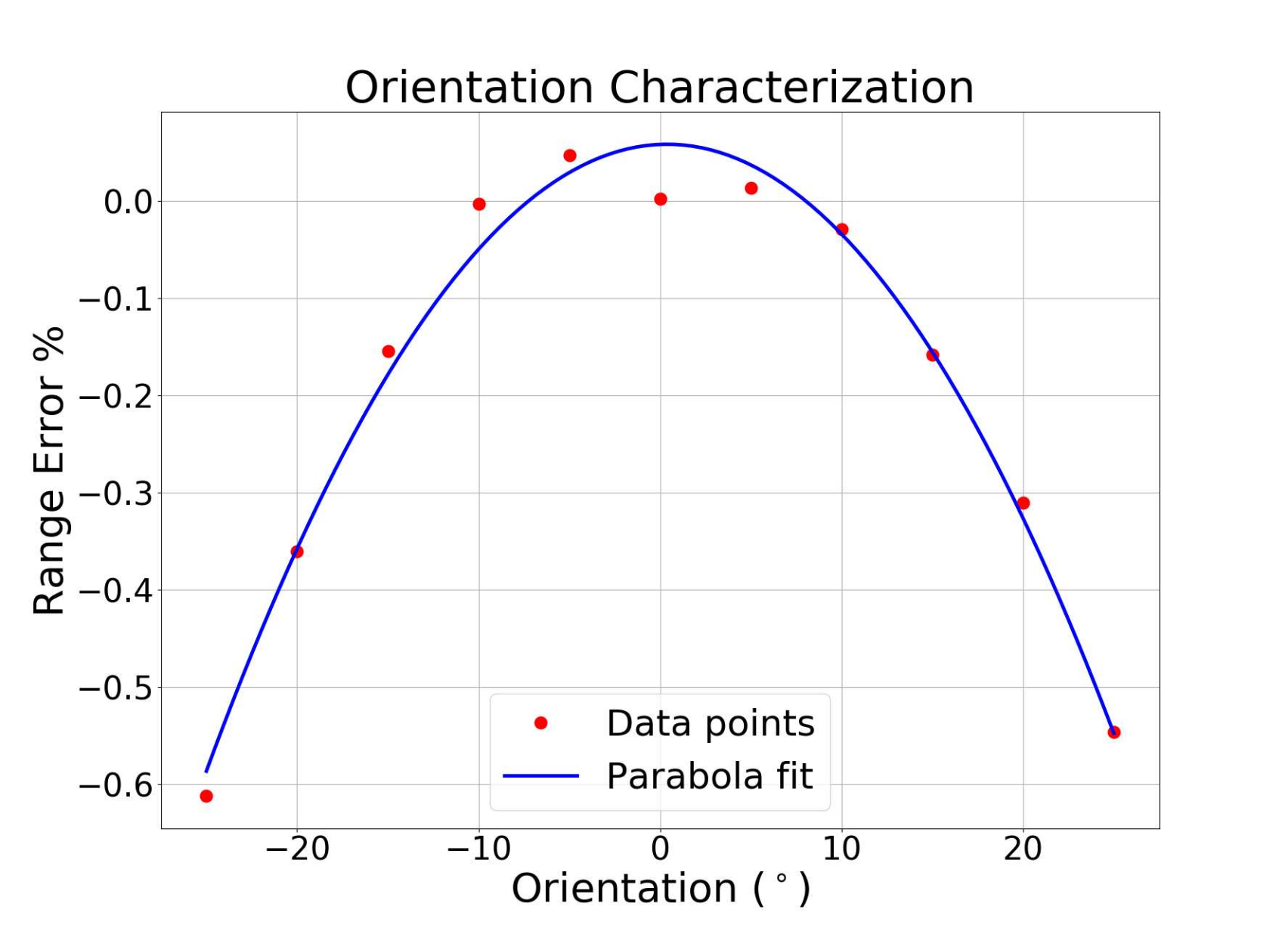}
    \caption{The figure shows the percentage error of the measurements on the distance estimation at different orientations along with the interpolation curve.}
    \label{fig:orientation_calibration}
\end{figure}

Similar to procedures performed for other sensors \cite{kneip2009characterization, cooper2018range}, we also characterized how the measurements are affected by the relative orientation between the \gls{tof} sensor and targets.
The robot's end-effector was kept at a constant distance from the target and commanded to move over a circular path to cover different orientations. 
The explored range was chosen such that all the beams intersected the target. During this procedure, the normal of the sensor intersected the reference point to the board as shown in \cref{fig:calibration setup}. 
Also in this experiment, both the relative error and the standard deviation are computed. 
The circular segment path had a radius of \SI{200}{\milli\meter} with respect to the reference point. The angle of the circular sector spanned from \SIrange{-25}{25}{\degree}, covered in \SI{5}{\degree} increments while the robot moved. 
This path is defined for the $x$-$y$ and $z$-$y$ planes.
Similar to the previous experiment, for each pose, we acquired and averaged 100 samples.

~\cref{fig:orientation_calibration} reports the recorded errors on the range estimation for the different orientations considered when moving in the $x$-$y$ plane. 
For this plot, range measurements were first compensated according to the range error observed in the previous characterization and then the error of the measured range with respect to the ground truth was computed. As can be seen from the plot, a parabolic curve interpolates well the results. In particular, a second-order polynomial was fitted to interpolate the data as follows:
\begin{equation}
    e_{xy} = -10^{-3}\theta^2 +7.78\times10^{-4}\theta + 0.06,
    \label{eq:e_xy}
\end{equation}
in which $\theta$ represents the orientation.

The experiments on the $y$-$z$ plane yielded similar results. Therefore, the error induced by the orientation of the \gls{tof} sensor with respect to the target can be approximated with the elliptic paraboloid:
\begin{equation}\label{eq:orientation}
  e= -10^{-3}(\theta^2 + \phi^2) +7.78\times10^{-4}\sqrt{(\theta^2 + \phi^2)}+ 0.06 . 
\end{equation}
In this case, $\theta$ is the angle in the $x$-$y$ plane 
as in (\ref{eq:e_xy}) and $\phi$ indicates the angle in the $y$-$z$ plane. This equation is used to correct the range measure and adopted for the sensor model as before.
Conversely, the standard deviation of the measure observed in these tests is not considered, as it was one order of magnitude smaller than the one observed in the range calibration (i.e., $<10^{-1}$ mm).

\section{Localization Experiments}\label{sec:validation}
\begin{figure}[t]
	\centering
	\subfigure[][]	{\label{fig:cf_press}\includegraphics[width=0.48\columnwidth]{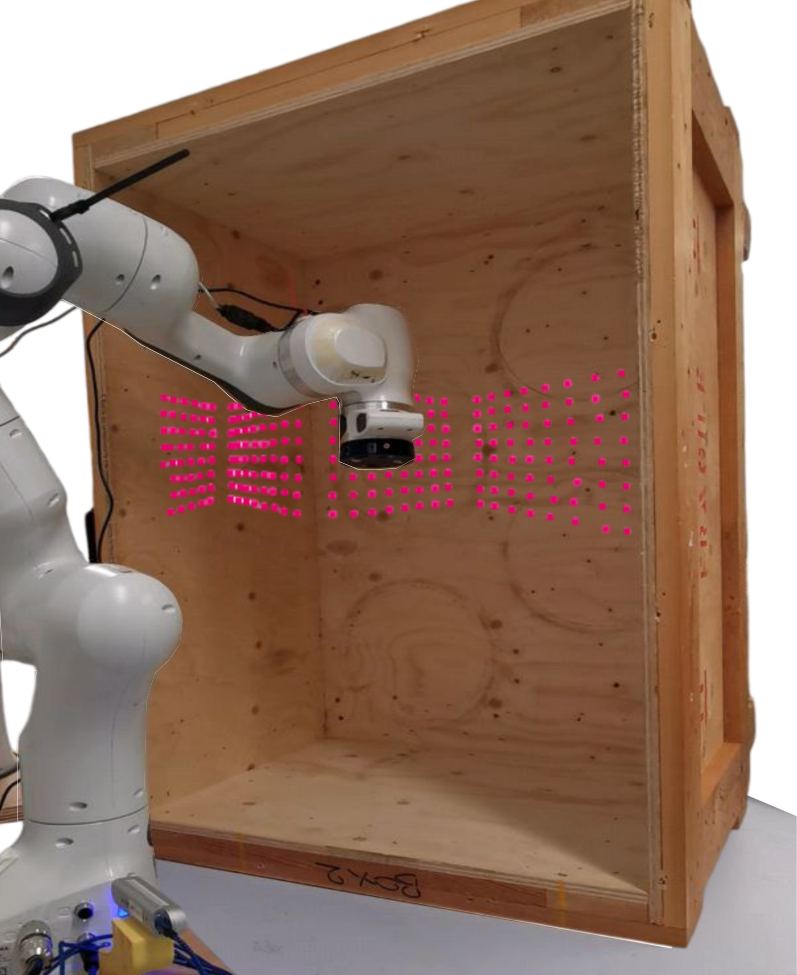}} 
	\subfigure[][]
	{\label{fig:cf_flow}\includegraphics[width=0.48\columnwidth]{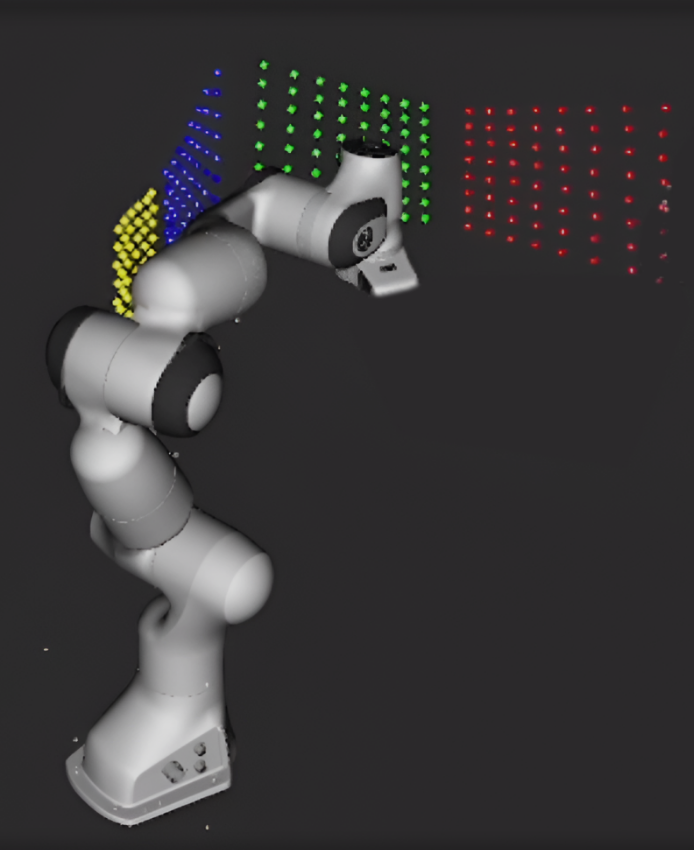}} 
	\caption{\gls{pc} reconstructed from \gls{tof} measurements at a given robot pose for the crate. (a) Real robot equipped with \gls{tof} sensors whose measurements are overlaid on the crate. (b) RViz visualization. A different color of the points is used to discriminate the different \gls{tof} unit generating the measurement.
  }
	\label{fig:localization_crate}
\end{figure}

\begin{figure}[]
	\centering
	\subfigure[][] 	{\label{fig:deer_real}\includegraphics[width=0.48\columnwidth]{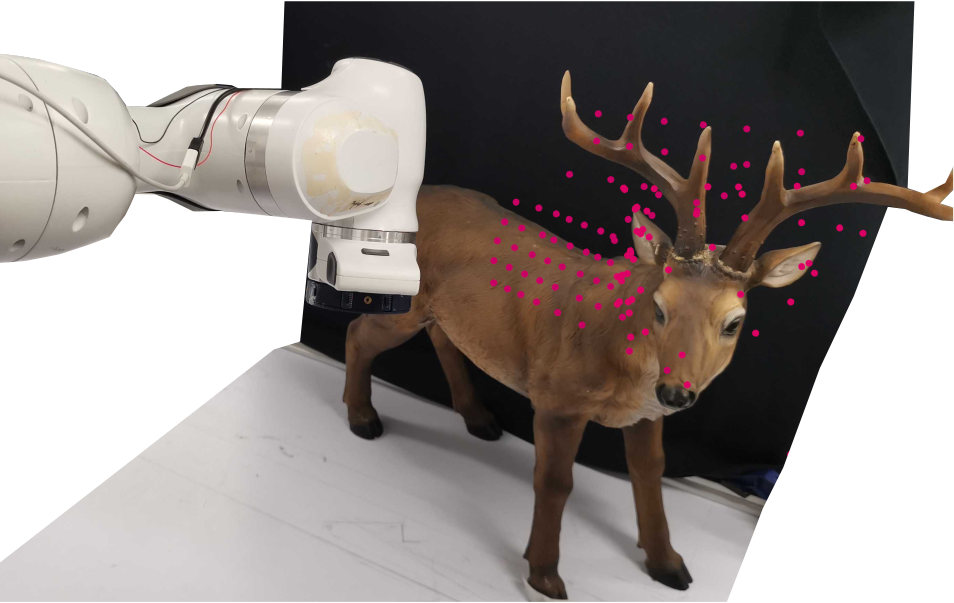}} 
	\subfigure[][]
	{\label{fig:deer_rviz}\includegraphics[width=0.48\columnwidth]{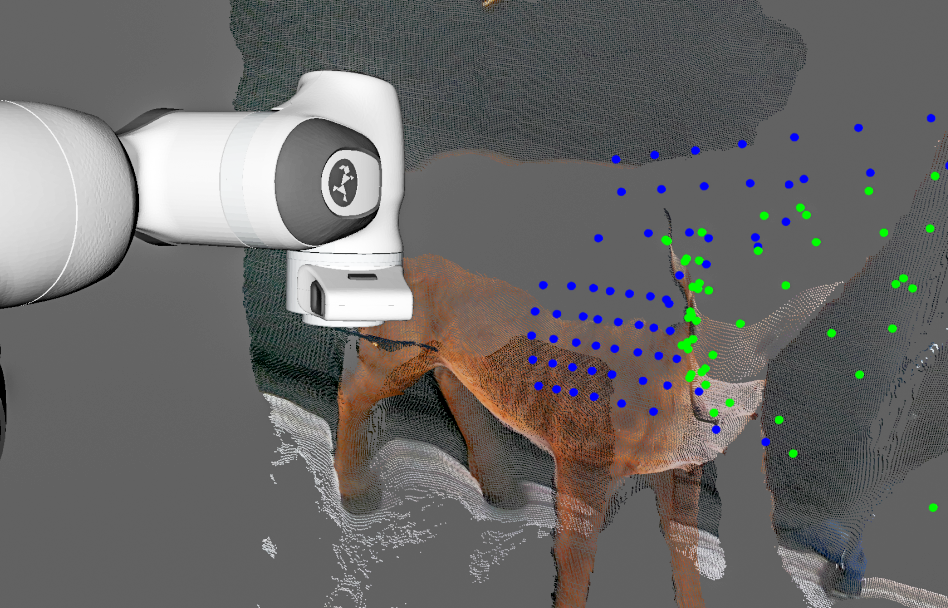}} 
	\caption{\gls{pc} reconstructed from \gls{tof} measurements at a given robot pose for the deer statue. (a) Overlay of \gls{tof} sensors measurements on the statue. (b) RViz visualization. The denser \gls{pc} for the deer was acquired using an Intel Realsense D415 depth camera mounted on the right side of the robot's base. 
  }
	\label{fig:localization_deer}
\end{figure}

The proposed model based on sensors' characterization -- which we will refer to as Probabilistic Sensor Model (PSM) -- is validated on two different object localization tasks, where we want to estimate the pose $x_o \in SE(2)$. As the objects lie on a flat surface, their height and out-of-plane orientation are fixed; we thus limit our study to position and orientation in the 2D $x$-$y$ plane without loss of generality.  \cref{fig:cf_press} and \cref{fig:deer_real} respectively show a crate and a statue of a deer placed in the robot workspace. With the crate, we reproduce a toy example of an inspection task for which the robot's end-effector is inside the space to explore. The crate presents strong features, e.g., corners and planes, that should be easy to detect with the \gls{tof}. Conversely, the deer statue allows us to test a more complex object with intricate features, which could be challenging due to the limited resolution of the sensors.

We benchmark PSM against two other methods. 
The first corresponds to the beam model in \cref{sec:methodology} built on the sensor datasheet -- referred to as DataSheet (DS). The second method -- Ideal Sensor (IS) -- assumes the \gls{tof} is free of uncertainties in its measurements.

\subsection{Data collection}\label{subsec:data_coll}

\gls{tof} data were acquired with 4 different sensors embedded in the same 3D-printed support as in \cref{sec:calibration}. 
Regarding the crate, throughout the entire data collection, its pose was kept stable and measured from the reference frame placed at the robot's base with measuring tape and a protractor. Since we know the crate shape, we can ray-cast every recorded sensor pose to measure the true distances between the sensor and the crate. Similarly, the statue was kept in a known pose for the data collection -- measured with an Intel RealSense D415 depth camera and an AprilTag tag~\cite{apriltag} -- while the mesh adopted for the ray-casting was adapted from the dense \gls{pc} collected in~\cite{border2024surface}.

For both objects, the robot was commanded in diverse poses -- 10 for the crate and 6 for the statue -- to collect measurements from the \gls{tof} sensors. An example of the output of the 4 sensors at a given robot pose is shown in \cref{fig:cf_press} in the form of a \gls{pc} for the crate. The deer statue only fell in two sensors' \gls{fov} given its size -- see~\cref{fig:deer_real}. In the following, we refer to \textit{data sample} to indicate the set of measurements collected at a given pose. 

The poses for the crate were chosen to ensure that most sensor beams fell on the object. Raw measurements corresponding to distances higher than \SI{600}{\milli\meter} were considered outside the crate volume and thus filtered out. Similarly, we surrounded the statue in a prismatic volume to reject the measurements.

It is worth noting that a more complete system may require a planner that moves the robot to the next position depending on the collected data~\cite{border2024surface} as well as a robust algorithm for scene segmentation. 
However, these aspects are out of the scope of this paper.
In these experiments, we are only interested in benchmarking the use of \gls{tof} in the localization task; therefore, we collected offline data samples in different robot poses to evaluate how performance in the localization task is affected by: (i) the number of \gls{tof} sensors; (ii) the number of data samples.

\begin{table*}[h!]
    \centering
    \caption{Crate Localization Results with respect to the number of acquisitions}
    \label{tab:one_tof}
    \setlength\extrarowheight{2pt}
    {\tiny
{
    \begin{tabular}{|c|c|c|c|c|c|c|c|c|c|c|}
    \hline
    \multirow{3}{*}{\textbf{Method}} & \multicolumn{10}{c|}{\textbf{Number of Data Samples}} \\ \cline{2-11} 
    & \multicolumn{2}{c|}{\textbf{2}} & \multicolumn{2}{c|}{\textbf{4}} & \multicolumn{2}{c|}{\textbf{6}} & \multicolumn{2}{c|}{\textbf{8}} & \multicolumn{2}{c|}{\textbf{10}} \\ \cline{2-11} 
    & $e_x$ (m) & $e_{\gamma}$$(^\circ)$ & $e_x$ (m)  & $e_{\gamma}$$(^\circ)$ & $e_x$ (m)  & $e_{\gamma}$$(^\circ)$ & $e_x$ (m)  & $e_{\gamma}$$(^\circ)$ & $e_x$ (m)  & $e_{\gamma}$$(^\circ)$ \\ \hline
    PSM (ours)                        & \cellcolor{cyan!10}0.031 ± 0.011 & \cellcolor{cyan!10}0.926 ± 0.630 & \cellcolor{cyan!10}0.049 ± 0.010 & \cellcolor{cyan!10}0.760 ± 0.280 & \cellcolor{cyan!10}0.040 ± 0.009 & \cellcolor{cyan!10}0.974 ± 0.520 & \cellcolor{cyan!10}0.055 ± 0.034 & \cellcolor{cyan!10}0.556 ± 0.454 & \cellcolor{cyan!10}0.049 ± 0.008 & \cellcolor{cyan!10}1.352 ± 0.183 \\ \hline
    DS                         &0.069 ± 0.004 &1.488 ± 0.194 &0.069 ± 0.005 &1.132 ± 0.312 &0.065 ± 0.004 &1.610 ± 0.273  &0.067 ± 0.004 &1.396 ± 0.402 &0.068 ± 0.011 &1.378 ± 0.713 \\ \hline
    IS                         &0.068 ± 0.003 &1.336 ± 0.171 &0.068 ± 0.005 &1.256 ± 0.476 &0.064 ± 0.008 &1.488 ± 0.317 &0.067 ± 0.005 &1.316 ± 0.345 &0.063 ± 0.006 &1.312 ± 0.489 \\ \hline
    \end{tabular}
    }}
\end{table*}

\begin{table*}[h!] 
    \centering
    \caption{Crate Localization Results with respect to the number of ToF sensors and the number of acquisitions}
    \label{tab:multiple_tof}
        \setlength\extrarowheight{2pt}
    {\tiny
    \resizebox{\textwidth}{!}{
    \begin{tabular}{|c|c|c|c|c|c|c|c|c|c|c|c|c|}
    \hline
    \multirow{3}{*}{\textbf{Method}} & \multicolumn{12}{c|}{\textbf{Number of ToFs / Number of Data Samples}} \\ \cline{2-13} 
                                & \multicolumn{2}{c|}{\textbf{2 / 4}} & \multicolumn{2}{c|}{\textbf{2 / 6}} & \multicolumn{2}{c|}{\textbf{3 / 4}} & \multicolumn{2}{c|}{\textbf{3 / 6}} & \multicolumn{2}{c|}{\textbf{4 / 4}}  & \multicolumn{2}{c|}{\textbf{4 / 6}} \\ \cline{2-13} 
                                & $e_x$ (m) & $e_{\gamma}$ $(^\circ)$ & $e_x$ (m) & $e_{\gamma}$ $(^\circ)$ & $e_x$ (m) & $e_{\gamma}$ $(^\circ)$ & $e_x$ (m) & $e_{\gamma}$ $(^\circ)$ & $e_x$ (m) & $e_{\gamma}$ $(^\circ)$ & $e_x$ (m) & $e_{\gamma}$ $(^\circ)$ \\ \hline
    PSM (ours)                       & \cellcolor{cyan!10}0.048 ± 0.012 & \cellcolor{cyan!10}0.522 ± 0.221 & \cellcolor{cyan!10}0.040 ± 0.016  &\cellcolor{cyan!10}0.370 ± 0.199  & \cellcolor{cyan!10}0.038 ± 0.014 & \cellcolor{cyan!10}0.298 ± 0.412 & \cellcolor{cyan!10}0.038 ± 0.014 & \cellcolor{cyan!10}0.798 ± 0.450 & \cellcolor{cyan!10}0.032 ± 0.009 & \cellcolor{cyan!10}0.412 ± 0.250 &  \cellcolor{cyan!10}0.036 ± 0.021 &  \cellcolor{cyan!10}0.888 ± 0.682 \\ \hline
    DS                         &0.068 ± 0.006 &1.688 ± 0.206 &0.071 ± 0.012  &1.094 ± 0.278  &0.071 ± 0.009 &1.444 ± 0.237 & 0.069 ± 0.008 &1.706 ± 0.348 &0.066 ± 0.005 &1.056 ± 0.221 &0.070 ± 0.008 &1.678 ± 0.381 \\ \hline
    IS                         &0.072 ± 0.006 &1.498 ± 0.329 &0.063 ± 0.010  &1.360 ± 0.437  &0.070 ± 0.008 &1.370 ± 0.145 &0.068 ± 0.004 &1.248 ± 0.550 &0.070 ± 0.007 &1.210 ± 0.270 &0.070 ± 0.006 &1.588 ± 0.224 \\ \hline
    \end{tabular}
    }}
\end{table*}

\begin{table*}[h!] 

    \centering
    \caption{Deer localization Results with respect to the number of ToF sensors and the number of acquisitions}
    \label{tab:deer_table}
        \setlength\extrarowheight{1pt}
    {\tiny
{
    \begin{tabular}{|c|c|c|c|c|c|c|c|c|}
    \hline
    \multirow{3}{*}{\textbf{Method}} & \multicolumn{8}{c|}{\textbf{Number of ToFs / Number of Data Samples}} \\ \cline{2-9} 
                                 & \multicolumn{2}{c|}{\textbf{1 / 4}} & \multicolumn{2}{c|}{\textbf{1 / 6}} & \multicolumn{2}{c|}{\textbf{2 / 4}} & \multicolumn{2}{c|}{\textbf{2 / 6}}   \\ \cline{2-9} 
                                & $e_x$ (m) & $e_{\gamma}$ $(^\circ)$ & $e_x$ (m) & $e_{\gamma}$ $(^\circ)$ & $e_x$ (m) & $e_{\gamma}$ $(^\circ)$ & $e_x$ (m) & $e_{\gamma}$ $(^\circ)$ \\ \hline
    PSM (ours)                        &\cellcolor{cyan!10}0.027 ± 0.004 &\cellcolor{cyan!10}1.232 ± 0.417 &\cellcolor{cyan!10}0.029 ± 0.008  &1.596 ± 0.593  & \cellcolor{cyan!10}0.033 ± 0.007 &1.585 ± 0.195 &0.033 ± 0.004 &\cellcolor{cyan!10}1.401 ± 0.399  \\ \hline
    DS                         &0.030 ± 0.006 &1.374 ± 0.424 &0.030 ± 0.008  &1.397 ± 0.317  &0.034 ± 0.010 &1.659 ± 0.094 &\cellcolor{cyan!10}0.029 ± 0.009 &1.612 ± 0.437 \\ \hline
    IS                         &0.027 ± 0.006 &1.362 ± 0.400 &0.033 ± 0.010  &\cellcolor{cyan!10}1.324 ± 0.385  &0.035 ± 0.006 &\cellcolor{cyan!10}1.449 ± 0.169 &0.030 ± 0.007 &1.658 ± 0.361  \\ \hline
    \end{tabular}
    }}
\end{table*}

\subsection{Experiment Details}

In line with what is presented in~\cref{sec:methodology}, a \gls{pf} approximates the belief of the 
object pose $x_o$ \cite{thrun2002probabilistic}. 

The generic particle is $x_t^j = \lbrace x, y, \gamma \rbrace$, where $\lbrace x, y \rbrace \in \mathbb{Re}^2$ represents the planar position of the reference frame rigidly attached to the object, and $\gamma \in \mathbb{Re}$ its orientation in the plane (as shown in~\cref{fig:calibration setup}).  

For each set of measurements $Z_t$ (see \cref{sec:methodology}), we compute its average $\bar{z}_t$ and correct its value by using (\ref{eq:range}) and (\ref{eq:orientation}) retrieved from the characterization experiments.
We also associate the standard deviation value for the given range and thus compute the likelihood $p(\bar{z}_t \mid x_t^j, o)$ for each particle $x_t^j$, i.e., the normal distribution centered at the true range value $z_t^*$. The true range is retrieved with a ray-casting technique between the known sensor pose and the 
object as devised by the particle. 
The likelihood value is then used to update the $j$-th weight. When more than one \gls{tof} is used, the associated likelihoods are obtained in the same way and then multiplied before updating the weights.

When considering the DS method, we do not correct the distance measure as there is no indication of this in the datasheet. Conversely, the standard deviation is provided as a function of the measured distance. Therefore, if the difference between $z_t^*$ and $\bar{z}_t$ for the $j$-th particle is smaller than the $\sigma$ provided by the datasheet, the likelihood is computed; otherwise, a null weight is given to the particle. As a matter of fact, the datasheet assumes that the measure follows a Gaussian distribution centered at the estimation itself, thus the likelihood will always be maximum or null depending on the distance between the measure and the true value.
Similarly, in the IS case, where the measure is assumed to be free from uncertainty, the measure belongs to a Gaussian centered in the measure itself, though with a very small deviation, i.e., in the order of the sensor resolution range (\SI{1}{\milli\meter}). In this case, if the distance between the estimation obtained with the ray-casting and the averaged sensor measure is below \SI{1}{\milli\meter}, the likelihood value is computed and the weight is updated with a non-zero value. 

\section{Results and Discussion}\label{sec:results}
Our analysis first considers a single \gls{tof} sensor and different numbers of data samples.
Then, for a fixed number of data samples, we analyze the effect of increasing the number of \gls{tof} sensors.
All \gls{pf} experiments initialize and draw 500 particles from a Gaussian distribution centered close to the ground truth and with an uncertainty of \SI{0.15}{\meter} and \SI{6}{\degree} for position and angle. We performed five localization trials for each method by re-initializing the \gls{pf}.
We compare each method by its absolute translation and rotation error -- $e_x$ and $e_\gamma$ -- defined as the norm of the difference between the true and estimated pose of the object. 

In~\cref{tab:one_tof} we report the trend of the mean errors and their standard deviation, related to the localization tests conducted with the crate when considering a single \gls{tof} sensor with respect to an increasing number of data samples. For each set of experiments, the lowest position and angular error are highlighted in light blue. 
The results demonstrate that PSM outperforms the other two methods, thus validating the characterization we performed. Specifically, across all experiments, PSM consistently delivered better performance, with a position error about \SI{0.02}{\meter} smaller on average. 
Increasing the number of data samples does not significantly affect localization performance. The same behavior is also observed for the two baseline methods. Although average error values slightly oscillate, PSM consistently delivers better performance. The oscillations in the position error \cref{tab:one_tof} are below \SI{10}{\milli\meter} in most of the cases, which is in line with the sensor noise we observed from the characterization experiments. 
A similar behavior is also observed in~\cref{tab:multiple_tof} when multiple sensors and different data samples are considered. Also, in this case, there are small oscillations in the errors while increasing the number of \gls{tof} sensors. However, as expected, PSM consistently outperforms the two baselines. It also appears that the number of \gls{tof} sensors does not significantly affect the localization accuracy.

With more samples -- whether collected by the same sensor at different times or jointly by multiple sensors -- the localization performance is expected to improve as more discriminative features are considered. Nevertheless, the crate we used in the localization experiment is a simple object that contains strong local features (planes and corners) that may be captured by a single \gls{tof} and with a small number of samples.
Conversely, the irregular shapes and fine details of the deer make ToF measurements challenging, thus limiting the effectiveness of using PSM. 
With fewer sensor beams intersecting the statue in each data sample (as shown in~\cref{fig:localization_deer}), fewer measurements are captured, reducing the impact of the PSM corrections.
As a result, while PSM generally outperforms the other methods, the difference in performance is less pronounced for the deer statue, with all three methods yielding similar results, i.e., $e_x < 0.035~m$ and $e_{\gamma} < 1.8^\circ$. 

It is also interesting to note that, across the different tests, there is little performance difference between the DS method (built on datasheet information) and the IS method, where no uncertainty is modeled. This motivates proper sensor characterization as discussed herein.

\section{Conclusion}\label{sec:conclusion}
In this paper, we propose a method to localize a target object in the manipulator workspace by using coarse information provided by \textit{tiny LiDARs}, i.e., the miniaturized VL53L5CX \gls{tof} sensors. We first characterize the sensors to assess how noise is affected by distance and orientation to the target. Then, we exploit the characterization curves and use an \gls{mcl} algorithm to estimate the pose of objects. The characterization not only provides information on the sensor's accuracy but also allows for an improvement in the localization task with respect to two baselines where measurements are considered with no uncertainty or with the one from the sensor's datasheet. 
The characterization in this work was done by considering just a white vinyl film wrapped around a flat board. Future work will focus on understanding how different physical properties of the target might affect the sensor readings. Additionally, the effects of environmental lighting and temperature should be further investigated.

In the proposed analysis, we localized a crate and a deer statue with respect to the robot base. In this context, we found that, in this setting, a single \gls{tof} sensor is sufficient to achieve the task. Future analyses will evaluate this aspect by considering different,
more complex objects as this may require a more sophisticated model for the sensor, taking into account other crucial sources of error, such as multipath interference or light scattering~\cite{jimenez2014modeling,may2009robust}. We aim to address these challenges in our future work.

\section*{Acknowledgment}\label{sec:acknowledgment}
The authors would like to thank Dr. Rowan Border (University of Cyprus) for sharing the Oxford Deer Point Cloud.

{\small
\bibliographystyle{IEEEtran}
\bibliography{ref}

\begin{thebibliography}{10}
\providecommand{\url}[1]{#1}
\csname url@samestyle\endcsname
\providecommand{\newblock}{\relax}
\providecommand{\bibinfo}[2]{#2}
\providecommand{\BIBentrySTDinterwordspacing}{\spaceskip=0pt\relax}
\providecommand{\BIBentryALTinterwordstretchfactor}{4}
\providecommand{\BIBentryALTinterwordspacing}{\spaceskip=\fontdimen2\font plus
\BIBentryALTinterwordstretchfactor\fontdimen3\font minus
  \fontdimen4\font\relax}
\providecommand{\BIBforeignlanguage}[2]{{%
\expandafter\ifx\csname l@#1\endcsname\relax
\typeout{** WARNING: IEEEtran.bst: No hyphenation pattern has been}%
\typeout{** loaded for the language `#1'. Using the pattern for}%
\typeout{** the default language instead.}%
\else
\language=\csname l@#1\endcsname
\fi
#2}}
\providecommand{\BIBdecl}{\relax}
\BIBdecl

\bibitem{zou2022}
Q.~Zou, Q.~Sun, L.~Chen, B.~Nie, and Q.~Li, ``A comparative analysis of lidar
  slam-based indoor navigation for autonomous vehicles,'' \emph{IEEE
  Transactions on Intelligent Transportation Systems}, vol.~23, no.~7, pp.
  6907--6921, 2022.

\bibitem{khan2021}
M.~U. Khan, S.~A.~A. Zaidi, A.~Ishtiaq, S.~U.~R. Bukhari, S.~Samer, and
  A.~Farman, ``A comparative survey of lidar-slam and lidar based sensor
  technologies,'' in \emph{2021 Mohammad Ali Jinnah University International
  Conference on Computing (MAJICC)}, 2021, pp. 1--8.

\bibitem{callenberg2021low}
C.~Callenberg, Z.~Shi, F.~Heide, and M.~B. Hullin, ``Low-cost spad sensing for
  non-line-of-sight tracking, material classification and depth imaging,''
  \emph{ACM Transactions on Graphics (TOG)}, vol.~40, no.~4, pp. 1--12, 2021.

\bibitem{ding2019proximity}
Y.~Ding, F.~Wilhelm, L.~Faulhammer, and U.~Thomas, ``With proximity servoing
  towards safe human-robot-interaction,'' in \emph{2019 IEEE/RSJ International
  Conference on Intelligent Robots and Systems (IROS)}.\hskip 1em plus 0.5em
  minus 0.4em\relax IEEE, 2019, pp. 4907--4912.

\bibitem{ding2020collision}
Y.~Ding and U.~Thomas, ``Collision avoidance with proximity servoing for
  redundant serial robot manipulators,'' in \emph{2020 IEEE International
  Conference on Robotics and Automation (ICRA)}.\hskip 1em plus 0.5em minus
  0.4em\relax IEEE, 2020, pp. 10\,249--10\,255.

\bibitem{caroleo}
G.~Caroleo, F.~Giovinazzo, A.~Albini, F.~Grella, G.~Cannata, and P.~Maiolino,
  ``A proxy-tactile reactive control for robots moving in clutter,'' in
  \emph{2024 IEEE/RSJ International Conference on Intelligent Robots and
  Systems (IROS)}, 2024, pp. 733--739.

\bibitem{al2020towards}
G.~A. Al, P.~Estrela, and U.~Martinez-Hernandez, ``Towards an intuitive
  human-robot interaction based on hand gesture recognition and proximity
  sensors,'' in \emph{2020 IEEE International Conference on Multisensor Fusion
  and Integration for Intelligent Systems (MFI)}.\hskip 1em plus 0.5em minus
  0.4em\relax IEEE, 2020, pp. 330--335.

\bibitem{yang2017pre}
B.~Yang, P.~Lancaster, and J.~R. Smith, ``Pre-touch sensing for sequential
  manipulation,'' in \emph{2017 IEEE International Conference on Robotics and
  Automation (ICRA)}.\hskip 1em plus 0.5em minus 0.4em\relax IEEE, 2017, pp.
  5088--5095.

\bibitem{sasaki2018robotic}
K.~Sasaki, K.~Koyama, A.~Ming, M.~Shimojo, R.~Plateaux, and J.-Y. Choley,
  ``Robotic grasping using proximity sensors for detecting both target object
  and support surface,'' in \emph{2018 IEEE/RSJ International Conference on
  Intelligent Robots and Systems (IROS)}.\hskip 1em plus 0.5em minus
  0.4em\relax IEEE, 2018, pp. 2925--2932.

\bibitem{koyama2018high}
K.~Koyama, M.~Shimojo, T.~Senoo, and M.~Ishikawa, ``High-speed high-precision
  proximity sensor for detection of tilt, distance, and contact,'' \emph{IEEE
  Robotics and Automation Letters}, vol.~3, no.~4, pp. 3224--3231, 2018.

\bibitem{ruget2022pixels2pose}
A.~Ruget, M.~Tyler, G.~Mora~Mart{\'\i}n, S.~Scholes, F.~Zhu, I.~Gyongy,
  B.~Hearn, S.~McLaughlin, A.~Halimi, and J.~Leach, ``Pixels2pose:
  Super-resolution time-of-flight imaging for 3d pose estimation,''
  \emph{Science Advances}, vol.~8, no.~48, p. eade0123, 2022.

\bibitem{hughes2018robotic}
D.~Hughes, J.~Lammie, and N.~Correll, ``A robotic skin for collision avoidance
  and affective touch recognition,'' \emph{IEEE Robotics and Automation
  Letters}, vol.~3, no.~3, pp. 1386--1393, 2018.

\bibitem{tsuji2021sensor}
S.~Tsuji and T.~Kohama, ``Sensor module combining time-of-flight with
  self-capacitance proximity and tactile sensors for robot,'' \emph{IEEE
  Sensors Journal}, vol.~22, no.~1, pp. 858--866, 2021.

\bibitem{caroleo2025soft}
G.~Caroleo, A.~Albini, and P.~Maiolino, ``Soft robot localization using
  distributed miniaturized time-of-flight sensors,'' in \emph{2025 IEEE 8th
  International Conference on Soft Robotics (RoboSoft)}, 2025, pp. 1--6.

\bibitem{liu2023multi}
X.~Liu, Y.~Li, Y.~Teng, H.~Bao, G.~Zhang, Y.~Zhang, and Z.~Cui, ``Multi-modal
  neural radiance field for monocular dense slam with a light-weight tof
  sensor,'' in \emph{Proceedings of the IEEE/CVF International Conference on
  Computer Vision}, 2023, pp. 1--11.

\bibitem{lancaster2022optical}
P.~Lancaster, P.~Gyawali, C.~Mavrogiannis, S.~S. Srinivasa, and J.~R. Smith,
  ``Optical proximity sensing for pose estimation during in-hand
  manipulation,'' in \emph{2022 IEEE/RSJ International Conference on
  Intelligent Robots and Systems (IROS)}.\hskip 1em plus 0.5em minus
  0.4em\relax IEEE, 2022, pp. 11\,818--11\,825.

\bibitem{THRUN200199}
\BIBentryALTinterwordspacing
S.~Thrun, D.~Fox, W.~Burgard, and F.~Dellaert, ``Robust monte carlo
  localization for mobile robots,'' \emph{Artificial Intelligence}, vol. 128,
  no.~1, pp. 99--141, 2001. [Online]. Available:
  \url{https://www.sciencedirect.com/science/article/pii/S0004370201000698}
\BIBentrySTDinterwordspacing

\bibitem{cooper2018range}
M.~A. Cooper, J.~F. Raquet, and R.~Patton, ``Range information characterization
  of the hokuyo ust-20lx lidar sensor,'' in \emph{Photonics}, vol.~5,
  no.~2.\hskip 1em plus 0.5em minus 0.4em\relax MDPI, 2018, p.~12.

\bibitem{kneip2009characterization}
L.~Kneip, F.~Tache, G.~Caprari, and R.~Siegwart, ``Characterization of the
  compact hokuyo urg-04lx 2d laser range scanner,'' in \emph{2009 IEEE
  International Conference on Robotics and Automation}.\hskip 1em plus 0.5em
  minus 0.4em\relax IEEE, 2009, pp. 1447--1454.

\bibitem{apriltag}
E.~Olson, ``Apriltag: A robust and flexible visual fiducial system,'' in
  \emph{2011 IEEE International Conference on Robotics and Automation}, 2011,
  pp. 3400--3407.

\bibitem{border2024surface}
R.~Border and J.~D. Gammell, ``The surface edge explorer (see): A
  measurement-direct approach to next best view planning,'' \emph{The
  International Journal of Robotics Research}, vol.~43, no.~10, pp. 1506--1532,
  2024.

\bibitem{thrun2002probabilistic}
S.~Thrun, ``Probabilistic robotics,'' \emph{Communications of the ACM},
  vol.~45, no.~3, pp. 52--57, 2002.

\bibitem{jimenez2014modeling}
D.~Jim{\'e}nez, D.~Pizarro, M.~Mazo, and S.~Palazuelos, ``Modeling and
  correction of multipath interference in time of flight cameras,'' \emph{Image
  and Vision Computing}, vol.~32, no.~1, pp. 1--13, 2014.

\bibitem{may2009robust}
S.~May, D.~Dr{\"o}schel, S.~Fuchs, D.~Holz, and A.~N{\"u}chter, ``Robust
  3d-mapping with time-of-flight cameras,'' in \emph{2009 IEEE/RSJ
  International Conference on Intelligent Robots and Systems}.\hskip 1em plus
  0.5em minus 0.4em\relax IEEE, 2009, pp. 1673--1678.

\end{thebibliography}
}

\end{document}